\documentclass[letterpaper]{article} 
\usepackage[preprint]{aaai2027}
\usepackage[hyphens]{url}  
\usepackage{graphicx} 
\usepackage{natbib}  
\usepackage{caption} 
\usepackage{algorithm}
\usepackage{algorithmic}
\usepackage{amsmath}
\usepackage{amssymb}
\usepackage{amsthm}

\usepackage[dvipsnames]{xcolor}

\usepackage{newfloat}
\usepackage{listings}
\DeclareCaptionStyle{ruled}{labelfont=normalfont,labelsep=colon,strut=off} 
\floatstyle{ruled}
\newfloat{listing}{tb}{lst}{}
\floatname{listing}{Listing}

\usepackage{booktabs}
\usepackage{multirow}
\usepackage{tabularx}
\usepackage[table]{xcolor}
\definecolor{tablegray}{gray}{0.94}
\definecolor{oursblue}{RGB}{232,242,252}
\title{HAFI-VLM: A Frequency Perspective for Diagnosing and Enhancing Visual Perception in Vision-Language Models}
\author {
    Jin Cui\textsuperscript{\rm 1}\equalcontrib,
    Chuanchang Su\textsuperscript{\rm 2}\equalcontrib,
    Jiayi Lu\textsuperscript{\rm 2},
    Xinyue Long\textsuperscript{\rm 1,\rm 3},
    Boran Zhao\textsuperscript{\rm 1,\rm 3}\corresponding,
    Pengju Ren\textsuperscript{\rm 1}
}
\affiliations {
    \textsuperscript{\rm 1}State Key Laboratory of Human-Machine Hybrid Augmented Intelligence,\\
    and Institute of Artificial Intelligence and Robotics, Xi'an Jiaotong University\\
    \textsuperscript{\rm 2}School of Computer Science and Technology, Xi'an Jiaotong University \\
    \textsuperscript{\rm 3}School of Software Engineering, Xi'an Jiaotong University \\
    andycui@stu.xjtu.edu.cn
}

\begin{document}

\maketitle

\begin{abstract}
Vision-language models (VLMs) remain unreliable when predictions require fine-grained visual evidence. We identify a previously overlooked cause: \emph{spectral response rigidity}. Despite substantial frequency variation across images and tasks, pretrained vision encoders exhibit persistent, encoder-specific layerwise spectral profiles that change only marginally under downstream fine-tuning. Since pretrained vision encoders only receive images, they cannot adapt spectral extraction to the evidence required by the current query. We therefore propose \textit{\textbf{HAFI-VLM}}, which introduces a task-conditioned frequency pathway while preserving the pretrained semantic representation. Hierarchical Adaptive Frequency Injection (\textit{\textbf{HAFI}}) retrieves complementary low-, mid-, and high-frequency evidence at multiple encoder depths using text-modulated, spatially aligned cross-attention. A Visual Enrichment Layer Adapter further recalibrates shallow LLM attention to effectively utilize the enriched visual tokens. Experiments on LLaVA-1.5 and Qwen2.5-VL demonstrate consistent improvements in general VQA, text-rich understanding, and hallucination robustness, outperforming representation-level enhancement methods and most resolution- or cropping-based approaches without additional high-resolution encoding. Mechanistic analyses show that \textit{\textbf{HAFI}} restores task-dependent spectral allocation while retaining semantic attention, establishing frequency enrichment as a distinct and effective route for improving VLM perception.
\end{abstract}


\section{Introduction}

Vision-language models (VLMs) have made substantial progress in connecting visual perception with language reasoning, yet remain unreliable when predictions depend on fine-grained evidence, such as small text, subtle attributes, object boundaries, and local textures. Existing solutions typically increase input resolution, aggregate multi-level features, or adapt the language backbone, implicitly assuming that the vision encoder can flexibly expose the evidence required by each downstream task. We revisit this assumption from a frequency-domain perspective by examining how pretrained vision encoders allocate spectral information across layers, images, and tasks.

Our analysis reveals a previously overlooked property, termed \emph{spectral response rigidity}. Despite substantial spectral variation across datasets and individual images, a given encoder produces nearly invariant layerwise frequency trajectories. Different encoders exhibit distinct profiles, indicating that spectral allocation is primarily determined by architecture and pretraining objectives rather than the current input. Effective rank and patch uniformity follow corresponding encoder-specific trajectories, indicating that the spectral response is embedded in the representation geometry established during pretraining. Moreover, this profile changes marginally under full downstream fine-tuning across optimization settings, demonstrating that conventional adaptation does not readily reorganize the pretrained spectral prior.

This rigidity is particularly restrictive in VLMs since standard vision encoders are conditioned only on the image. The same image therefore yields identical visual tokens whether the query concerns global semantics, fine-grained textual regions, or subtle attributes. The limitation is not a universal lack of high-frequency information, but the inability to adapt spectral extraction to query-specific evidence requirements. When relevant evidence lies in a frequency band attenuated by the pretrained profile, the language backbone receives insufficient visual evidence and may instead rely on linguistic priors, impairing OCR, fine-grained recognition, localization, and hallucination robustness.

To address this limitation, we propose \textbf{\textit{Hierarchical Adaptive Frequency Injection}} (\textbf{\textit{HAFI}}), which introduces a query-conditioned spectral retrieval pathway into the vision encoder. A parallel DCT branch represents the image as low-, mid-, and high-frequency tokens, while lightweight cross-attention modules retrieve complementary evidence at multiple encoder depths. The pretrained visual representation is preserved through residual injection, avoiding destructive modification of the semantic capabilities established during pretraining, and independent modules accommodate the distinct filtering behavior of different stages. A text-conditioned band gate further selects frequency components according to the current query, enabling different questions about the same image to induce different spectral allocations.

\begin{figure*}[t]
    \centering
    \includegraphics[width=1.0\textwidth]{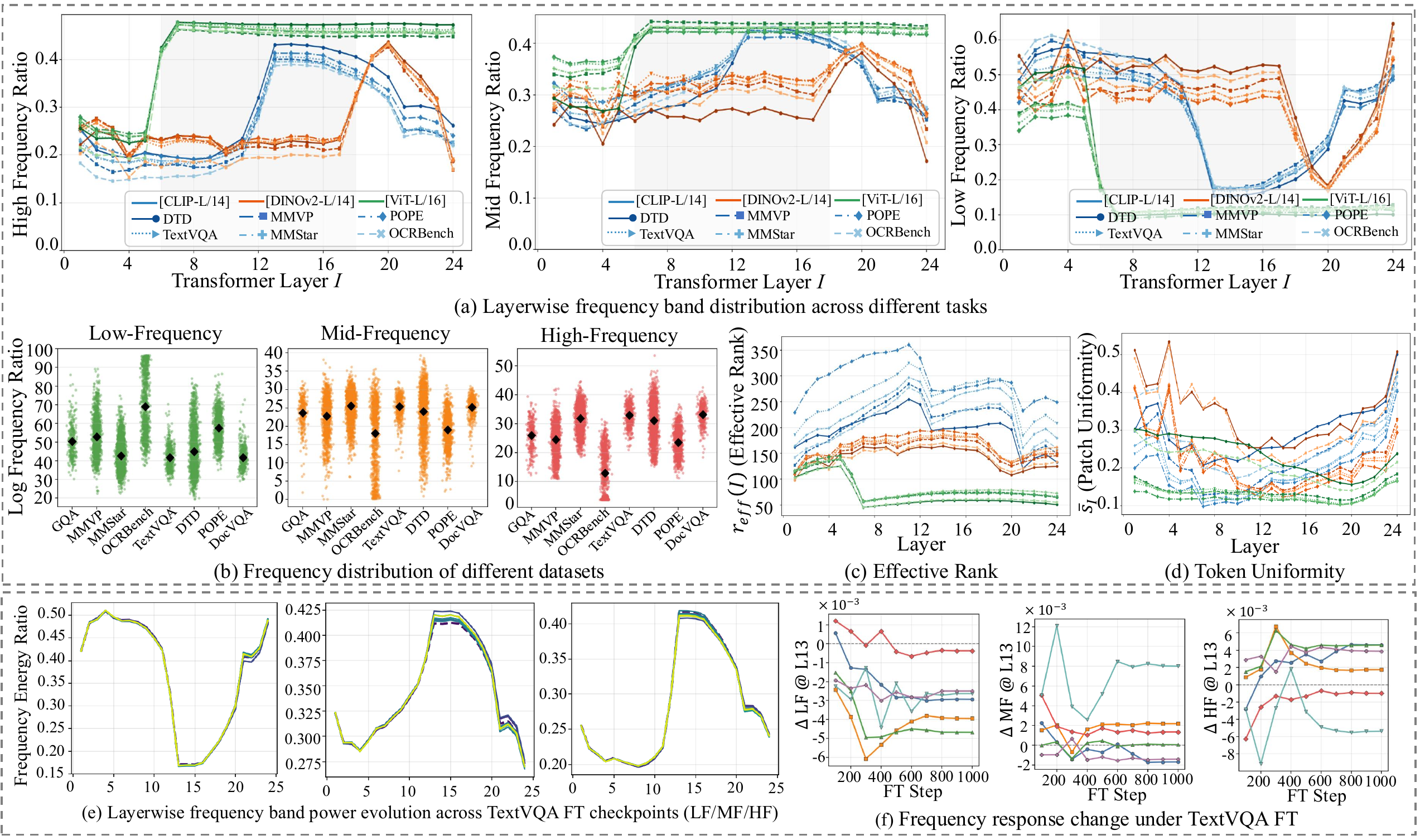}
    \caption{\textbf{Spectral response rigidity of pretrained vision encoders.} Although the benchmarks exhibit substantially different input spectra in (b), the layerwise frequency trajectories remain highly consistent within each encoder in (a). (c--d) Effective rank and patch uniformity reveal encoder-specific representation compression that co-evolves with the spectral profiles. (e--f) Across TextVQA fine-tuning checkpoints, evaluation datasets, and optimization steps, the layerwise band allocation changes only marginally, indicating that standard downstream adaptation does not readily reconfigure the pretrained spectral response.}
    \vspace{-2mm}
    \label{fig:spectral_analysis}
    \vspace{-2mm}
\end{figure*}

Encoder-side enrichment also changes the distribution of visual tokens presented to the pretrained language backbone. We therefore introduce a \textit{\textbf{Visual Enrichment Layer Adapter}} (\textbf{\textit{VEL-Adapter}}), which efficiently adapts the query and key projections of shallow LLM layers responsible for visual-information transfer. Together, \textbf{\textit{HAFI}} and VEL-Adapter form \textit{\textbf{HAFI-VLM}}, jointly improving task-conditioned evidence extraction and its subsequent utilization. Experiments across VLM families demonstrate consistent gains in general VQA, text-rich understanding, and hallucination robustness. Mechanistic analyses further show that \textbf{\textit{HAFI}} restores task-dependent spectral allocation while preserving attention to semantically relevant regions, establishing frequency enrichment as a distinct and complementary route for improving VLM perception. Our main contributions are summarized:
\begin{itemize}
   \item We identify \emph{spectral response rigidity} in pretrained vision encoders and reveal a structural mismatch between visual encoding and query-dependent evidence requirements. 

    \item We propose \textit{\textbf{HAFI-VLM}}, which preserves the pretrained semantics while enabling task-conditioned spectral evidence retrieval and effective utilization by the LLM.

    \item Experiments show consistent improvements in general and text-rich understanding and robustness. Analyses confirm the spectral allocation varies with task semantics, suggesting that \textbf{\textit{HAFI}} supplements fine-grained evidence without discarding the pretrained semantic pathway.
\end{itemize}

\begin{figure*}
    \centering
    \includegraphics[width=1\linewidth]{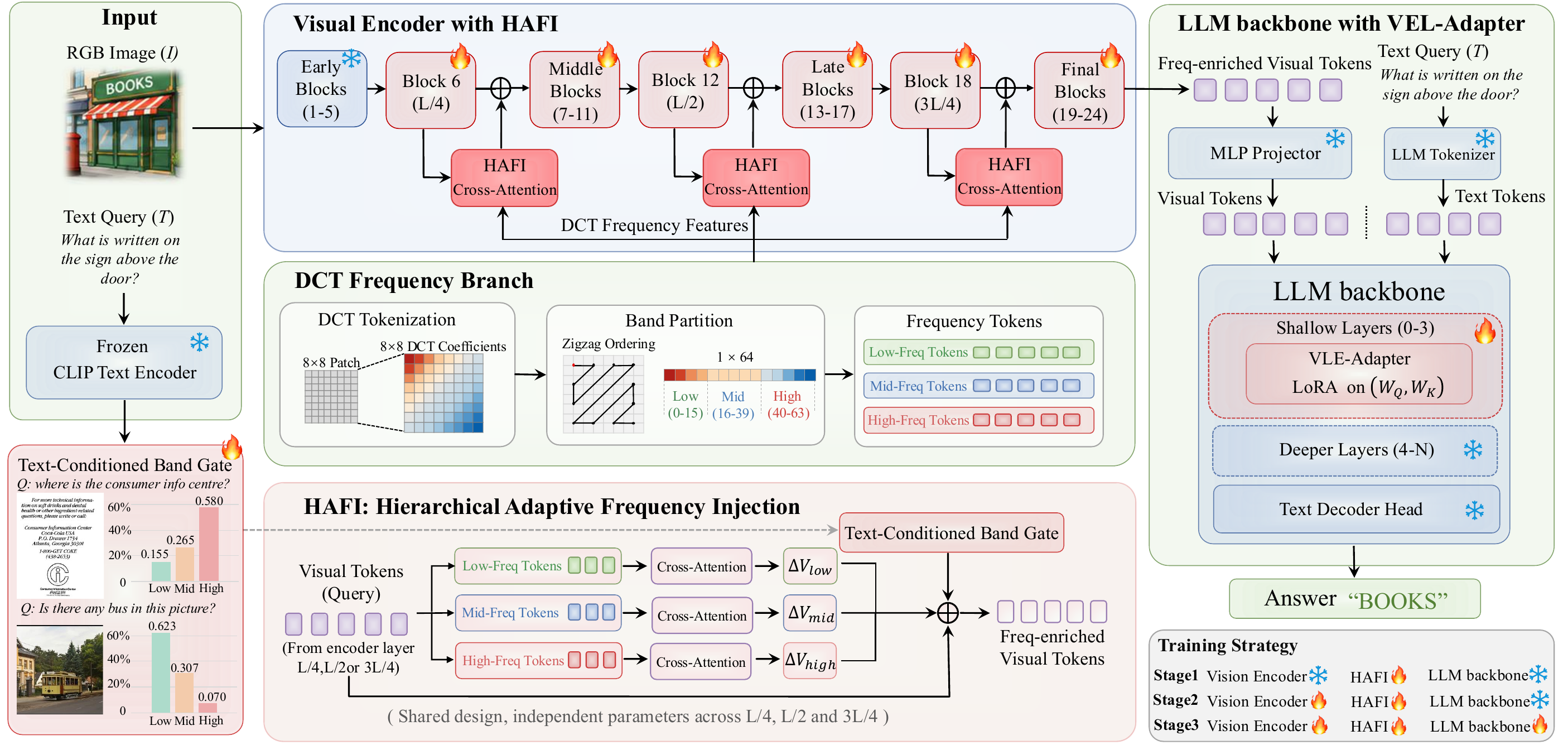}
    \caption{\textbf{Overview of \textbf{\textit{HAFI-VLM}}.} A DCT branch constructs multi-frequency tokens, which are retrieved and residually injected through text-conditioned \textbf{\textit{HAFI}} modules at multiple vision encoder depths. VEL-Adapter adapts the shallow visual--language interface in the LLM backbone to the frequency-enriched visual tokens.}
    \label{fig:main-fig}
    \vspace{-3mm}
\end{figure*}

\section{Related Work}

\paragraph{Query-Aware Visual Adaptation in MLLMs.}
Input-level methods recover fine-grained evidence through multi-scale encoding, regional cropping, or adaptive resolution \cite{Shi2024WhenDW,Luo2024FeastYE,Zhang2025MLLMsKW,shi2025sdrpn,yang2025visionthink,lin2026adaptvision,shi2026qzoom}. Readout-level approaches instead fuse features from multiple encoder layers or introduce instruction-aware visual queries \cite{Dai2023InstructBLIPTG,Yao2024DenseCF,Cao2024MMFuserMM}. TGIF further makes layer fusion prompt-dependent, but conditions the selection of already-computed encoder outputs \cite{Lin2026TextGuidedLF}. SteerViT, LIVE, and SCVM move language- or state-conditioned modulation into the encoding process, primarily operating on spatial or semantic representations \cite{ruthardt2026steerable,mao2026live,liu2026scvm}. Complementary to these approaches, \textbf{\textit{HAFI}} conditions the retrieval and multi-depth injection of explicit spectral evidence, operating at native resolution without repeated passes through the pretrained vision encoder.

\paragraph{Frequency-Domain Modeling and Spectral Adaptation.}
FcaNet formulates channel attention as multi-spectral DCT compression, while SpectFormer combines spectral operators with self-attention \cite{qin2021fcanet,patro2025spectformer}. In VLMs, DocPedia represents high-resolution documents in the DCT domain, whereas Fourier Compressor exploits frequency redundancy for visual-token compression \cite{feng2024docpedia,wang2025fouriercompressor}. Beyond architectural design, recent work analyzes how spectral information evolves across encoder layers, and FGINet progressively injects frequency cues into a pretrained backbone for a specialized vision task \cite{kitessa2026spectral,zhou2026fginet}. These studies reveal depth-dependent spectral transformations and the utility of frequency-semantic fusion, but leave open how a general-purpose VLM should adapt its spectral evidence to different language queries over the same image. HAFI targets this setting by retrieving band-separated frequency tokens through text-conditioned gates at multiple encoder depths while retaining the pretrained semantic pathway; a lightweight VEL-Adapter further aligns shallow LLM attention with the enriched visual representation.

\section{Frequency-Domain Analysis}
\label{sec:frequency_analysis}

\paragraph{Layerwise spectral probing.}
We analyze three representative vision encoders---CLIP ViT-L/14, DINOv2 ViT-L/14, and supervised ViT-L/16---on six benchmarks covering texture recognition, OCR, hallucination evaluation, and general visual reasoning.
For an image $I$, let $X_{\ell}(I)\in\mathbb{R}^{N\times D}$ denote the patch-token representation at layer $\ell$.
After rearranging the tokens into their spatial grid, we compute the channel-averaged spectral power $\bar{P}_{\ell}^{I}(u,v)$ and define the normalized energy ratio of frequency band $b\in\{\mathrm{L},\mathrm{M},\mathrm{H}\}$ as

\begin{equation}
\rho_{\ell}^{b}(I)
=
\frac{
\sum_{(u,v)\in\Omega_b}\bar{P}_{\ell}^{I}(u,v)
}{
\sum_{u,v}\bar{P}_{\ell}^{I}(u,v)
},
\label{eq:band_ratio}
\end{equation}

where $\Omega_b$ denotes the corresponding spectral region.
We additionally report the effective rank of the patch-token matrix to measure the concentration of the singular-value spectrum and the mean pairwise cosine similarity between patch tokens (termed patch uniformity) to measure the average angular similarity. These quantities characterize representation geometry independently of the bandwise energy ratios.

\paragraph{Spectral response rigidity.}
Despite substantial variation in input spectra across datasets and samples (Fig.~\ref{fig:spectral_analysis}(b)), each encoder produces highly consistent layerwise frequency trajectories across benchmarks (Fig.~\ref{fig:spectral_analysis}(a)). By contrast, CLIP, DINOv2, and supervised ViT exhibit markedly different profiles, indicating that spectral evolution is governed primarily by architecture and pretraining rather than the current input. Effective rank and patch uniformity follow corresponding encoder-specific trajectories (Fig.~\ref{fig:spectral_analysis}(c,d)), showing that spectral allocation forms part of the representation geometry established during pretraining. We term this persistent behavior \emph{spectral response rigidity}. This rigidity is closely coupled with the semantic capability acquired during pretraining.  The pretraining objective organizes visual features into semantic directions while learning invariances to visual variations that are less useful for its supervision. It reflects an encoder-specific semantic--spectral trade-off: each pretraining objective preserves or compresses visual variations according to the semantic abstractions it favors, producing a stable spectral prior.

\paragraph{Persistence under full downstream fine-tuning.}
To determine whether this rigidity results from frozen parameters, we fully unfreeze the vision encoder and fine-tune it on TextVQA under multiple optimization settings. The layerwise trajectories remain close to their pretrained profiles throughout training and evaluation (Fig.~\ref{fig:spectral_analysis}(e)), while bandwise changes remain small and show no systematic redistribution of spectral energy (Fig.~\ref{fig:spectral_analysis}(f)). Downstream fine-tuning therefore adapts how existing visual representation is used, but leaves its pretrained spectral allocation largely intact.

\paragraph{Consequences for vision-language modeling.}
Spectral response rigidity is particularly restrictive in VLMs because a standard vision encoder is conditioned only on the image. For two questions $Q_1$ and $Q_2$ concerning the same image $I$,

\begin{equation}
\Phi_v(I,Q_1)
=
\Phi_v(I,Q_2)
=
\Phi_v(I),
\label{eq:query_independent_encoder}
\end{equation}

where $\Phi_v$ denotes the vision encoder. The encoder therefore produces the same visual tokens regardless of whether the query concerns global scene structure, small text, object boundaries, or subtle local attributes.

Such queries require different combinations of spectral evidence, yet an image-only encoder cannot reallocate capacity toward the relevant frequency bands. When task-critical components have been attenuated or compressed, the projector and language model receive an insufficient visual representation that language reasoning cannot recover, increasing perception failures and reliance on linguistic priors.

The central limitation is therefore the absence of \emph{task-conditioned spectral extraction}: pretrained vision encoders apply a persistent spectral allocation even though different queries require different visual evidence. One last question is whether the frequency adaptation is necessary for the language backbone and will eventually benefit the VLMs. We will verify this through our following experiments.


\section{Method}
\label{sec:method}

\begin{table*}[t]
\centering
\caption{Comparison with representative vision-enhanced MLLMs grouped by backbone family. Results are taken from the original papers or official model cards when available or reproduced according to the specifications in their paper.}
\label{tab:main_results}

\small
\setlength{\tabcolsep}{2pt}
\renewcommand{\arraystretch}{1.10}

\begin{tabular}{llcccccccc}
\toprule

\textbf{Model}
& \textbf{Visual Design}
& \textbf{VQAv2}
& \textbf{GQA}
& \textbf{SQA-I}
& \textbf{TextVQA}
& \textbf{DocVQA}
& \textbf{ChartQA}
& \textbf{MMB}
& \textbf{POPE}
\\

\midrule



\rowcolor{gray!18}
LLaVA-1.5-7B~\cite{Liu2023ImprovedBW}
&CLIP ViT-L/14@336 baseline &78.5&62.0&66.8&58.2&21.5&18.1& 64.3&86.6
\\

\rowcolor{yellow!13}
Dense Connector~\cite{Yao2024DenseCF}
&Multi-layer visual feature fusion&79.5&63.8&69.5&59.2&24.8 &21.5&66.8&86.6
\\

\rowcolor{yellow!13}
MMFuser~\cite{Cao2024MMFuserMM} & Fuse multiple encoder layers
&79.1&62.8&68.7&58.8&--&--&67.5&86.3
\\

\rowcolor{yellow!13}
TGIF~\cite{Lin2026TextGuidedLF}
&Reweight encoder layer features&62.6&70.1&59.0&--&--&66.4&69.9&--
\\

\rowcolor{yellow!13}
$S^2$~\cite{Shi2024WhenDW} & Multi-scale feature extraction
&80.0&59.4&63.8&52.6&27.1&18.9&64.3&87.9
\\

\rowcolor{green!10}
ViCrop~\cite{Zhang2025MLLMsKW} & Attention-guided RoI cropping
&76.5&61.0&--&57.2&27.0&20.0&--&88.6
\\

\rowcolor{green!10}
SD-RPN~\cite{shi2025sdrpn}& Self-distilled region refinement &79.4&62.7&69.6&58.7&33.9&20.1&64.3&87.8
\\

\rowcolor{green!10}
LLaVA-HR~\cite{Luo2024FeastYE} & Dual-resolution dual-encoder
&81.9&64.2&65.1&67.1&\textbf{45.2}&24.0&68.0&88.0
\\

\rowcolor{green!10}
Q-Zoom~\cite{shi2026qzoom}& Query-aware RoI zooming 
&--&--&--&66.5&34.2&20.6&--&--
\\

\rowcolor{blue!10}
Finetuned-LLaVA& Finetuned on the same settings 
&79.3&63.6&65.1&64.3&24.9&18.9&64.3&86.6
\\

\rowcolor{blue!10}
\textbf{HAFI-VLM-LLaVA}
&HAFI + VEL-Adapter
&\textbf{82.8}
&\textbf{67.9}
&\textbf{70.3}
&\textbf{68.5}
&31.4
&\textbf{24.6}
&\textbf{71.5}
&\textbf{89.0}
\\

\midrule



\rowcolor{gray!18}
Qwen2.5-VL-7B~\cite{Bai2025Qwen25VLTR} & Native ViT baseline
&77.1&72.1&83.1&84.9&95.1&79.8&84.1&86.7
\\

\rowcolor{yellow!13}
SD-RPN~\cite{shi2025sdrpn}& Self-distilled region refinement 
&81.5&65.6&87.3&83.5&93.6&85.5&84.1& 83.1
\\

\rowcolor{yellow!13}
AdaptVision~\cite{lin2026adaptvision}& Adaptive visual acquisition &83.0&61.1&\textbf{88.8}&85.2&92.6&75.9&84.2& 86.8
\\

\rowcolor{green!10}
VisionThink~\cite{yang2025visionthink}& Query-based resolution routing &83.3&61.3&88.7&85.4&94.4&79.8&83.1&86.0
\\

\rowcolor{green!10}
Q-Zoom~\cite{shi2026qzoom}& Query-aware RoI zooming 
&84.0&61.9&87.2&83.5&94.3&\textbf{85.6}&84.0&87.1
\\

\rowcolor{blue!10}
Finetuned-Qwen& Finetuned on the same settings 
&80.3&72.0&83.5&85.2&95.1&80.2&84.2&86.6
\\

\rowcolor{blue!10}
\textbf{HAFI-VLM-Qwen}
&HAFI + VEL-Adapter
&\textbf{83.7}
&\textbf{72.9}
&86.6
&\textbf{86.2}
&\textbf{96.1}
&83.9
&\textbf{85.0}
&\textbf{87.9}
\\

\bottomrule
\end{tabular}

\end{table*}

\vspace{-6mm}


\subsection{Overall Architecture}
\label{sec:overall_architecture}

Let $\Phi_v=\{\phi_i\}_{i=1}^{L}$ denote the CLIP visual encoder, $\Pi$ the frozen multimodal projector, and $\Phi_l$ the LLM. Given an image $I$ and a text query $T$, the visual stream starts from the standard CLIP patch embedding,
\begin{equation}
    V_0=\operatorname{PatchEmbed}(I), \qquad
    V_0 \in \mathbb{R}^{B \times N_v \times D}.
\end{equation}
In parallel, the frequency branch maps the same image into low-, mid-, and high-frequency token sets:
\begin{equation}
    F=\Psi_f(I)=\{F^{\mathrm{low}},F^{\mathrm{mid}},F^{\mathrm{high}}\}.
\end{equation}
These frequency tokens are not concatenated to the input sequence. Instead, they are retrieved after selected visual encoder blocks,
$
    \mathcal{S}_{\mathrm{inj}}
    =
    \{L/4,L/2,3L/4\}.
$
For the $i$-th encoder block, we first compute $V_i^-=\phi_i(V_{i-1})$. If $i\notin\mathcal{S}_{\mathrm{inj}}$, we set $V_i=V_i^-$. Otherwise, \textbf{\textit{HAFI}} applies a frequency update conditioned on $V_i^-$, $F$, and $T$. After the final visual layer, the frequency-enriched tokens are projected and decoded as
\begin{equation}
    Z_v=\Pi(V_L), \qquad
    p_\Theta(y\mid I,T)=\Phi_l(y\mid Z_v,T),
\end{equation}
where $\Theta$ denotes the stage-dependent trainable parameters.

\subsection{DCT Frequency Tokenization}
\label{sec:dct_tokenization}

The frequency branch provides an explicit spectral basis complementary to the persistent allocation learned by the pretrained vision encoder.
We first de-normalize the RGB image, convert it to YCrCb space, partition it into non-overlapping $8\times8$ patches, and apply a two-dimensional DCT to each patch.
The resulting $64$ coefficients are flattened in zigzag order and partitioned into three bands:$\Omega_{\mathrm{low}}  = \{0,\ldots,15\},
\Omega_{\mathrm{mid}}  = \{16,\ldots,39\},
\Omega_{\mathrm{high}} = \{40,\ldots,63\}.$


The complete frequency branch can be written as

\vspace{-3mm}

\begin{equation}
\begin{aligned}
I_{\mathrm{ycc}}
&=
\operatorname{RGB2YCrCb}
\bigl(
\operatorname{Denorm}(I)
\bigr),\\
\bar{Z}^{b}
&=
\operatorname{Pool}_{2\times2}
\left[
\operatorname{Split}_{\Omega_b}
\left(
\operatorname{Zigzag}
\left(
\operatorname{DCT}_{8\times8}(I_{\mathrm{ycc}})
\right)
\right)
\right],\\
F^{b}
&=
\operatorname{LN}_{b}
\left(
W_f^{b}\bar{Z}^{b}
+
e_{\mathrm{pos}}
+
e_{\mathrm{band}}^{b}
\right),
\qquad
b\in\mathcal{B},
\end{aligned}
\label{eq:dct_tokens}
\end{equation}

where
$\mathcal{B}=\{\mathrm{low},\mathrm{mid},\mathrm{high}\}$.
The band-specific projection $W_f^b$ maps coefficients into the visual hidden dimension, while
$e_{\mathrm{pos}}$ and $e_{\mathrm{band}}^b$ preserve their spatial location and frequency identity. For a $336\times336$ image, the initial $42\times42$ DCT-patch grid is reduced to $21\times21$, yielding $N_f=441$ tokens per band.
\textit{\textbf{HAFI}} does not require $N_f=N_v$, since the two grids interact through cross-attention.

\subsection{Hierarchical Adaptive Frequency Injection}
\label{sec:hafi}

The spectral response of the pretrained encoder varies substantially with depth.
We therefore use an independent \textit{\textbf{HAFI}} module at each
$i\in\mathcal{S}_{\mathrm{inj}}$,
allowing different stages to retrieve distinct frequency evidence.
Within each module, visual tokens provide spatially resolved queries, while the text representation introduces task-semantic guidance.

\paragraph{Residual text modulation.}
We obtain a global text representation from a frozen CLIP text encoder,

\begin{equation}
t
=
\operatorname{Pool}
\left(
E_{\mathrm{CLIP}}^{\mathrm{text}}(T)
\right), \qquad v_i=\operatorname{Pool}(V_i^-)
\label{eq:text_representation}
\end{equation}

At inject layer $i$, the visual query and text modulation are projected into a shared bottleneck space of dimension $D_s$:

\vspace{-3mm}

\begin{equation}
Q_i=P_{Q,i}\left(\operatorname{LN}(V_i^-)\right)
+\eta_i\,\mathbf{1}_{N_v}(P_{T,i}\left(\operatorname{LN}(t)\right))^{\top}.
\end{equation}


Here, $q_i^t=P_{T,i}\left(\operatorname{LN}(t)
\right)$ is broadcast across the visual-token dimension and acts as a residual semantic bias rather than replacing the image-dependent queries.
The scalar
$\eta_i=\eta_{\max}\sigma(c_i)$
is sigmoid-bounded and initialized near zero, enabling training begins from the original image-conditioned retrieval behavior.
The text projection is shared by the three frequency bands within each injection layer, introducing only one lightweight semantic modulator per depth.

\paragraph{Spatially aligned band-wise retrieval.}
The frequency tokens are first mapped to the same bottleneck dimension,

\begin{equation}
R_i^{b}
=
P_{KV,i}(F^{b}),
\qquad
b\in\mathcal{B},
\label{eq:frequency_projection}
\end{equation}

where $P_{KV,i}$ is shared across bands within layer $i$.
Since the visual and frequency branches use different spatial grids, unconstrained attention must learn their geometric correspondence entirely from downstream supervision.
We instead introduce a lightweight spatial alignment prior.

Let
$p_n^{v}\in[0,1]^2$
and
$p_m^{f}\in[0,1]^2$
denote the normalized center coordinates of the $n$-th visual patch and the $m$-th frequency patch, respectively.
For band $b$, we define

\vspace{-3mm}

\begin{equation}
B_{i,nm}^{b}
=
-\tau_i^{b}
\left\|
p_n^{v}-p_m^{f}
\right\|_2^2, 
\tau_i^{b}
=
\operatorname{softplus}
\left(
\widehat{\tau}_i^{b}
\right).
\label{eq:spatial_prior}
\end{equation}

The prior softly favors frequency evidence originating from nearby image regions while retaining global retrieval when supported by content similarity. Only one non-negative scalar $\tau_i^b$ is introduced for each layer--band pair and is shared across attention heads.

For the $h$-th attention head, the spatially biased retrieval is

\vspace{-3mm}

\begin{equation} \small
A_{i,h}^{b}
=
\operatorname{Softmax}_{m}
\left(
\frac{
Q_{i,h}(K_{i,h}^{b})^{\top}
}{
\sqrt{d_h}
}
+
B_i^{b}
\right),
O_{i,h}^{b}
=
A_{i,h}^{b}U_{i,h}^{b}
\label{eq:spatial_cross_attention}
\end{equation}

where
$K_{i,h}^{b}$ and $U_{i,h}^{b}$
are the key and value projections of $R_i^b$.
The band-specific update is then

\vspace{-3mm}

\begin{equation}
\Delta V_i^{b}
=
P_{\uparrow,i}
\left[
\operatorname{Concat}_{h}
\left(
O_{i,h}^{b}
\right)
W_{O,i}^{b}
\right].
\label{eq:band_update}
\end{equation}

The down- and up-projections are shared across bands within each layer, while the compact attention output projections remain band-specific. This provides sufficient frequency-specific capacity without duplicating the full adaptation pathway.

\paragraph{Text-conditioned band allocation.}
Different tasks may require different combinations of spectral evidence. We therefore use a layer-level gate conditioned on both the current visual state and the same text representation used in Eq.~\eqref{eq:text_representation}:

\vspace{-3mm}

\begin{equation}
g_i
=
\operatorname{Softmax}
\left(
\operatorname{MLP}_{\mathrm{gate},i}
([v_i;t])
\right)
\in\mathbb{R}^{B\times3}
\label{eq:band_gate}
\end{equation}

The resulting weights are broadcast over visual tokens and combine the three updates. HAFI updates the pretrained visual state through



\vspace{-3mm}

\begin{equation}
V_i
=
V_i^-
+
\lambda_i \sum_{b\in\mathcal{B}}
g_i^{b}\Delta V_i^{b}
,
\qquad
\lambda_i
=
\lambda_{\max}\sigma(a_i).
\label{eq:hafi_update}
\end{equation}

\vspace{-3mm}


\begin{figure*}
    \centering
    \includegraphics[width=1\linewidth]{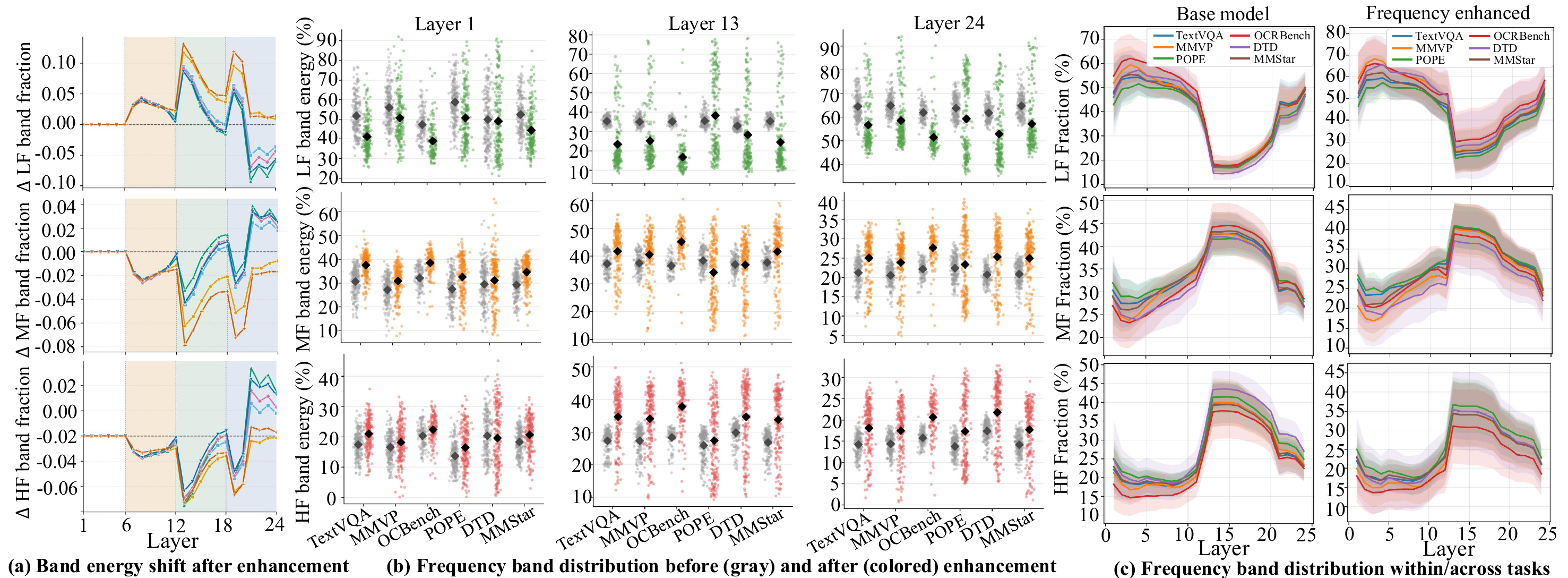}
    \caption{\textbf{Task-adaptive spectral rebalancing induced by HAFI.} (a) HAFI produces structured, layer-dependent shifts across frequency bands. (b) Frequency enhancement yields more diverse sample-level and cross-task band distributions than the base model. (c) Frequency enhancement didn't alter the frequency response characteristics established during pre-training; instead, it amplifies the sample-level variance in spectral utilization, as well as the task-dependent frequency utilization across benchmarks.}
    \label{fig:frequency_band_energy}
    \vspace{-4mm}
\end{figure*}

\subsection{Visual Enrichment Layer Adapter}
\label{sec:vel_adapter}

Frequency injection enriches the visual tokens but also changes the representation encountered by the language backbone. A fully frozen decoder may therefore fail to retrieve the newly exposed evidence, particularly in the shallow layers where visual information is transferred into language tokens.

VEL-Adapter applies LoRA only to the query and key projections of the first four LLM layers $\mathcal{S}_{\mathrm{vel}} = \{0,1,2,3\}$,

\vspace{-3mm}

\begin{equation}
W_{m}^{(\ell)}
\leftarrow
W_{m}^{(\ell)}
+
B_{m}^{(\ell)}A_{m}^{(\ell)},
m\in
\{
\mathrm{q\_proj},
\mathrm{k\_proj}
\},
\ell\in\mathcal{S}_{\mathrm{vel}}.
\label{eq:vel_lora}
\end{equation}

Adaptations recalibrate how language tokens match enriched visual evidence without modifying the transferred value content. All value projections, feed-forward networks, normalization layers, and deeper decoder blocks remain frozen, preserving pretrained language reasoning pathway.

\begin{figure}
    \centering
    \includegraphics[width=1\linewidth]{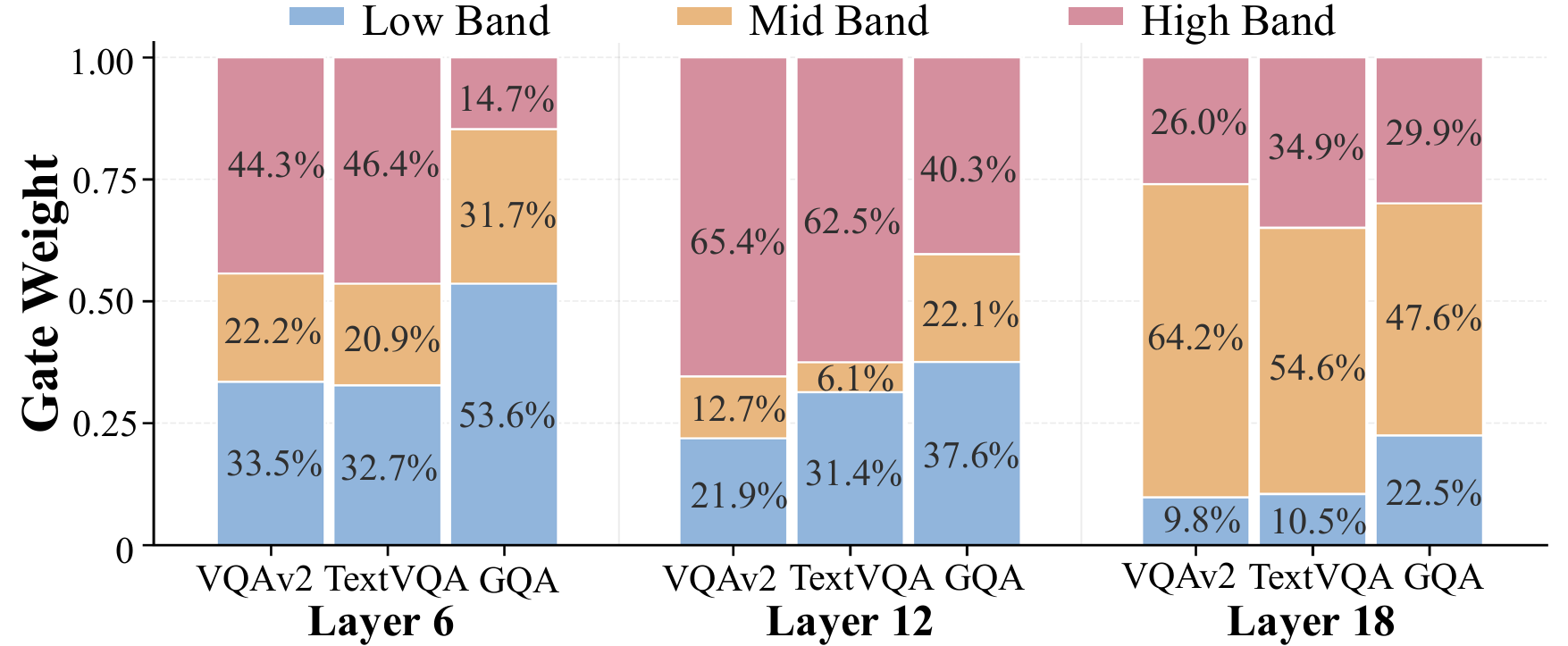}
    \caption{\textbf{Text-conditioned band allocation across tasks and depths.} The learned frequency injection weights vary across different benchmarks and across injection layers, demonstrating task- and depth-dependent spectral selection.}
    \label{fig:ablation-gate}
    \vspace{-2mm}
\end{figure}

\subsection{Training Objective and Optimization}
\label{sec:training}

The primary objective is the standard autoregressive language-modeling loss,

\vspace{-3mm}

\begin{equation}
\mathcal{L}_{\mathrm{task}}
=
-
\sum_{t=1}^{T_y}
\log
p_{\Theta}
\left(
y_t
\mid
y_{<t},I,T
\right).
\label{eq:task_loss}
\end{equation}

To initialize the frequency branch with meaningful spectral representations, we employ a DCT-space reconstruction loss,

\vspace{-3mm}

\begin{equation}
\begin{gathered}
\mathcal{L}_{\mathrm{rec}}^{\mathrm{dct}}
=
\sum_{b\in\mathcal{B}}
\frac{w_b}{B N_f 3 k_b}
\left\|
\hat{Z}^{b}-\bar{Z}^{b}
\right\|_2^2,\\
w_{\mathrm{low}}:w_{\mathrm{mid}}:w_{\mathrm{high}}
=
1:2:4.
\end{gathered}
\end{equation}

We additionally use a weak batch-level gate-balancing regularizer during joint adaptation.

\vspace{-4mm}

\begin{equation} \small
    \bar{g}^{b}
    =
    \frac{1}{B|\mathcal{S}_{\mathrm{inj}}|}
    \sum_{n=1}^{B}
    \sum_{i\in\mathcal{S}_{\mathrm{inj}}}
    g_{i,n}^{b}, \quad
    \mathcal{L}_{\mathrm{bal}}
    =
    \sum_{b\in\mathcal{B}}
    \bar{g}^{b}
    \log(\bar{g}^{b}+\epsilon).
\end{equation}


It discourages global collapse to a single band without imposing uniform allocation on individual samples.

Training follows four stages. \textbf{Stage 0} pretrains only the frequency encoder and reconstruction head using $\mathcal{L}_{\mathrm{rec}}^{\mathrm{dct}}$.
\textbf{Stage 1} freezes the pretrained VLM and frequency branch and warms up the complete \textbf{\textit{HAFI}} stack, including the text modulators and parameters, using $\mathcal{L}_{\mathrm{task}}$. \textbf{Stage 2} jointly optimizes \textbf{\textit{HAFI}}, the frequency encoder, selected visual self-attention projections, and the shallow VEL-Adapter parameters with


\vspace{-2mm}

\begin{equation}
\mathcal{L}^{(2)}
=
\mathcal{L}_{\mathrm{task}}
+
\beta_{\mathrm{rec}}
\mathcal{L}_{\mathrm{rec}}^{\mathrm{dct}}
+
\beta_{\mathrm{bal}}
\rho(t)
\mathcal{L}_{\mathrm{bal}},
\label{eq:joint_loss}
\end{equation}

where
$\beta_{\mathrm{rec}}=0.1$,
$\beta_{\mathrm{bal}}=0.01$, and
$\rho(t)$ linearly activates the balancing term during the final $30\%$ of this stage.
\textbf{Stage 3} removes all auxiliary objectives and fine-tunes only \textbf{\textit{HAFI}} and VEL-Adapter using
$\mathcal{L}_{\mathrm{task}}$.


\begin{figure}
    \centering
    \includegraphics[width=1\linewidth]{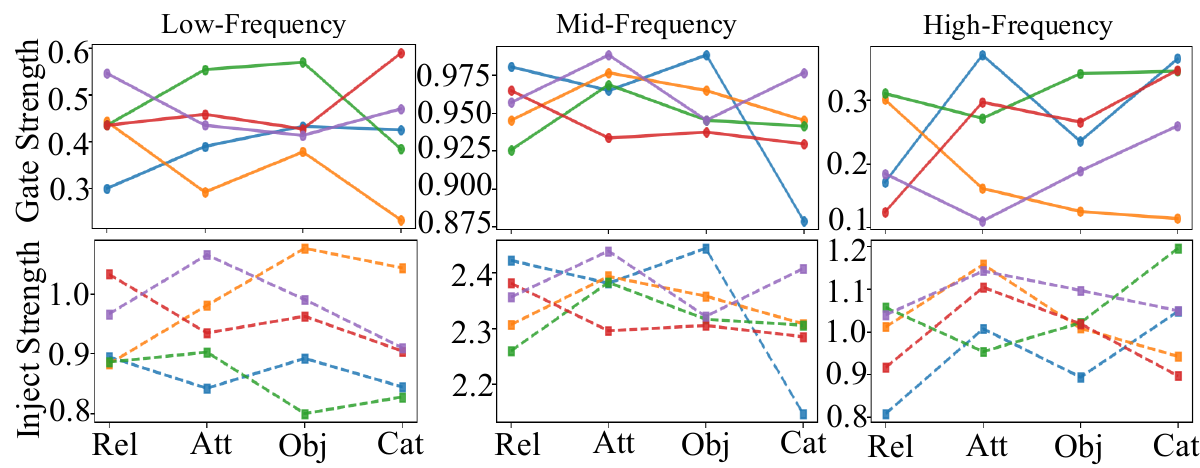}
    \caption{\textbf{Text semantic guided frequency injection.} Gate and injection strengths for different frequency bands vary across different queries, showing that different textual task types induce distinct frequency-injection patterns.}
    \label{fig:semantic}
    \vspace{-4mm}
\end{figure}

\section{Experiments}

\begin{table}[t]
\centering
\caption{Comparison with representative vision-enhanced MLLMs grouped by backbone family. Results are taken from the original papers or official model cards.}
\label{tab:component_ablation}

\small
\setlength{\tabcolsep}{3pt}
\renewcommand{\arraystretch}{1.10}

\begin{tabular}{lcccc}
\toprule

\textbf{Variant}
& \textbf{VQAv2}
& \textbf{TextVQA}
& \textbf{ChartQA}
& \textbf{POPE}
\\

\midrule


\rowcolor{gray!18}
LLaVA-1.5-7B&78.5&58.2&18.1&86.6
\\

\rowcolor{blue!10}
\textbf{HAFI-VLM-LLaVA}
&\textbf{82.8}
&\textbf{68.5}
&\textbf{24.6}
&\textbf{89.0}
\\
\midrule
w./o VEL-Adapter
&76.3$_{\downarrow\,6.5}$
&58.9$_{\downarrow\,9.3}$
&17.2$_{\downarrow\,7.4}$
&85.9$_{\downarrow\,7.9}$
\\

Early 6 layers LoRa
&82.9$_{\textcolor{blue}{\uparrow\,0.1}}$
&69.0$_{\textcolor{blue}{\uparrow\,0.5}}$
&24.7$_{\textcolor{blue}{\uparrow\,0.1}}$
&88.9$_{\downarrow\,0.1}$
\\

Unfreeze all
&82.2$_{\downarrow\,0.6}$
&67.0$_{\downarrow\,1.5}$
&24.5$_{\downarrow\,0.1}$
&88.2$_{\downarrow\,0.8}$
\\

\midrule

w./o Text-condition
&81.7$_{\downarrow\,1.1}$
&67.6$_{\downarrow\,0.9}$
&24.2$_{\downarrow\,0.4}$
&86.9$_{\downarrow\,2.1}$
\\

\midrule

Shared HAFI module
&81.0$_{\downarrow\,1.8}$
&66.9$_{\downarrow\,1.6}$
&24.2$_{\downarrow\,0.4}$
&86.9$_{\downarrow\,2.1}$
\\





\bottomrule
\end{tabular}

\end{table}

\subsection{Experimental Setup}


\paragraph{Benchmarks and Baselines.}
We evaluate \textit{\textbf{HAFI-VLM}} on six benchmarks grouped as in Table~\ref{tab:main_results}. We compare \textit{\textbf{HAFI-VLM}} with two representative families of visual enhancement methods: 
(1) representation-level approaches that fuse, reweight, or refine multi-layer visual features, and (2) input-level approaches that acquire additional evidence through higher resolution, regional cropping, or RoI zooming.

\paragraph{Implementation Details.}
We instantiate \textit{\textbf{HAFI-VLM}} on two representative backbones: LLaVA-1.5-7B and Qwen2.5-VL-7B. \textit{\textbf{HAFI}} is inserted at layers $\{6,12,18\}$ for LLaVA and $\{8,16,24\}$ for Qwen2.5-VL, while both variants use the same frequency-injection architecture and a frozen CLIP ViT-L/14 text encoder for text-conditioned band gating. We train the models on 8 NVIDIA A100 GPUs. Full hyperparameters and dataset mixture are provided in Appendix~\ref{app:training_details}.


\subsection{Main results}
As shown in Table~\ref{tab:main_results}, \textit{\textbf{HAFI-VLM}} achieves the strongest performance on fine-grained VQA and consistently surpasses representation-level methods on text-rich understanding. It also outperforms most high-resolution and region-based approaches while using the native resolution and a single encoder pass, yielding a favorable accuracy--efficiency trade-off. This demonstrates that task-conditioned frequency injection exposes discriminative visual evidence that cannot be recovered by recombining the incomplete semantic features. Its strong POPE performance suggests that query-relevant spectral evidence provides a more complete perceptual basis for reasoning and reduces reliance on linguistic priors.

Figure~\ref{fig:frequency_band_energy} shows that HAFI increases the task-specific \emph{allocability} of spectral information while preserving the pretrained layerwise profile. Compared with the base encoder, the enhanced model exhibits substantially broader cross-sample coverage in frequency bands, with the dispersion approaching the diversity in the input images shown in Fig.~\ref{fig:spectral_analysis}(b). Clear inter-task differences also emerge: fine-grained and text-rich benchmarks reduce low-frequency dominance and emphasize mid- and high-frequency components.  Figure~\ref{fig:frequency_band_energy}(c) confirms that the narrow cross-sample distributions of the base encoder are broadened into task-differentiated spectral responses. Meanwhile, the overall layerwise trajectories remain close to the pretrained profile. Stronger injection induces larger spectral shifts but degrades performance, indicating that effective enrichment should increase spectral variation without disrupting pretrained semantic--spectral prior.

Figures~\ref{fig:ablation-gate} and~\ref{fig:semantic} further reveal that this allocation varies with encoder depth and task semantics. Shallow layers retrieve the three bands relatively evenly, intermediate layers emphasize high-frequency evidence, and deeper layers favor mid-frequency components for semantic integration. At the query level, attribute- and category-oriented questions induce stronger mid- and high-frequency injection, consistent with their reliance on edges, textures, and appearance cues, whereas relation- and object-oriented questions favor low- and mid-frequency structure. Spatial visualizations in the Appendix further show that \textit{\textbf{HAFI}} preserves pretrained semantic attention while retrieving frequency evidence from query-relevant regions. \textit{\textbf{HAFI}} therefore performs semantic-aware spectral retrieval rather than indiscriminate detail amplification, establishing frequency enrichment as an effective and complementary route to improving VLM perception.

\subsection{Ablation Studies}

\begin{figure}
    \centering
    \includegraphics[width=1\linewidth]{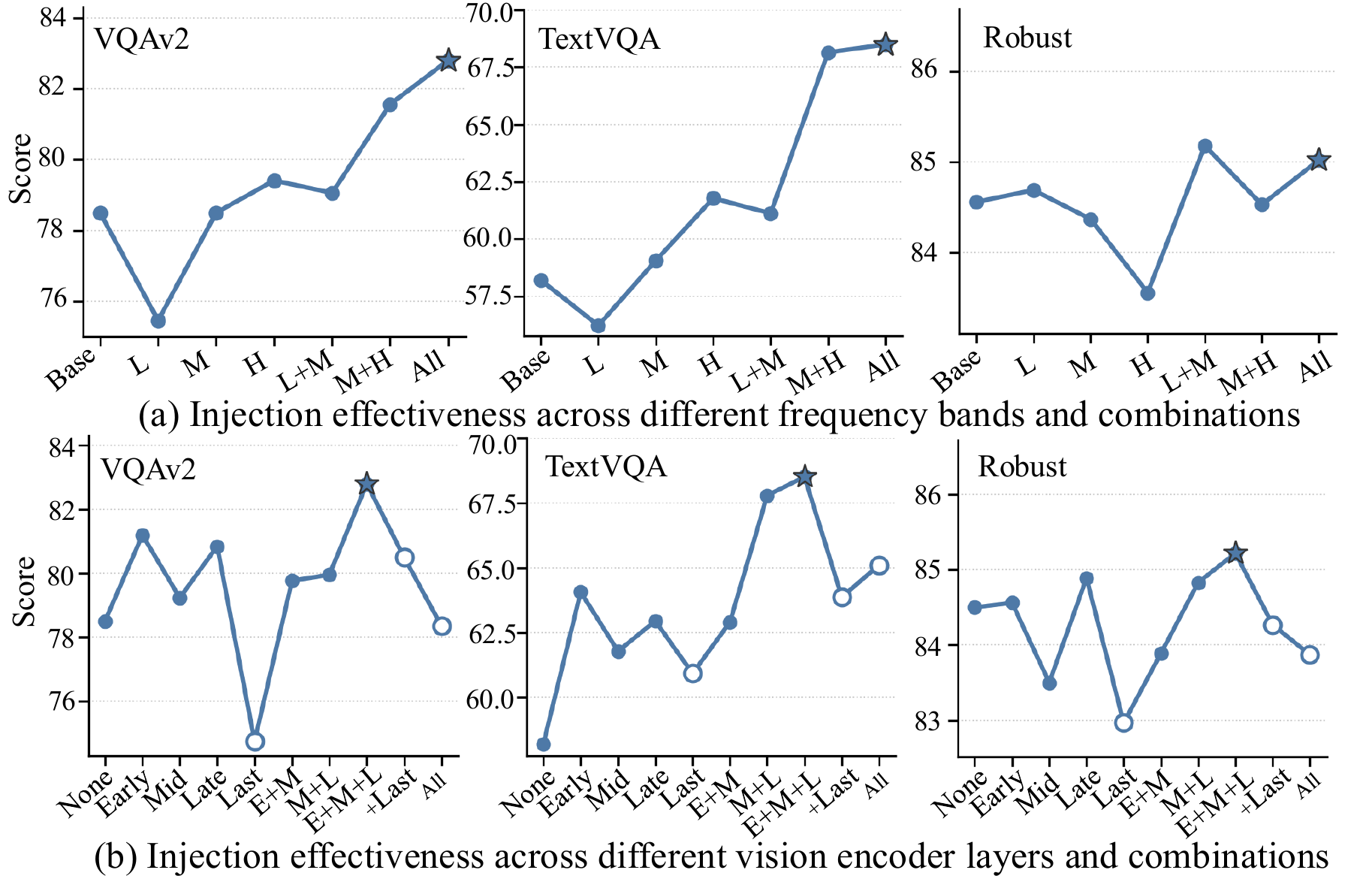}
    \caption{Ablations on frequency bands and injection depths.}
    \label{fig:frequency_layer_ablation}
    \vspace{-4mm}
\end{figure}

\paragraph{Frequency bands and injection depths.}
Figure~\ref{fig:frequency_layer_ablation}(a) shows that jointly injecting all three frequency bands performs best. Low-frequency-only injection generally degrades accuracy, suggesting that dominant coarse structure encourages shortcuts while providing limited complementary evidence. Mid- and high-frequency cues are more beneficial, but neither matches their combination, highlighting the importance of preserving complementary information across the spectrum.  Figure~\ref{fig:frequency_layer_ablation}(b) further favors sparse multi-depth injection. This agrees with the functional specialization in Figure~\ref{fig:ablation-gate}: different stages retrieve distinct evidence, whereas excessive injection interferes with the pretrained semantic pathway. Variations of the injection layers within the same depth range produce only minor differences, making multi-depth coverage more important than the exact layer choice.

\paragraph{Role of VEL-Adapter.}
Removing VEL-Adapter reduces performance by an average of $7.8$ points across the four benchmarks in Table~\ref{tab:component_ablation}, showing that a frozen language backbone cannot effectively utilize frequency-enriched visual tokens. Frequency injection shifts the visual-token distribution and perturbs the pretrained visual--language alignment established during pretraining. Adapting only the shallow LLM layers restores this interface, while fully unfreezing the backbone remains inferior, supporting localized rather than indiscriminate language-model adaptation.


\paragraph{Text conditioning and layer-specific injection.}
Removing text conditioning consistently degrades performance, confirming that query semantics are essential for retrieving task-relevant spectral evidence. Combined with the spatial prior, text modulation turns generic frequency amplification into targeted perceptual retrieval. Sharing \textit{\textbf{HAFI}} parameters is also suboptimal, indicating that early, intermediate, and late encoder stages require distinct retrieval functions due to their different levels of visual and semantic abstraction.

\section{Conclusion}

We reveal \textit{spectral response rigidity} in pretrained vision encoders, producing largely query-invariant frequency allocations despite task demands. We thus propose \textit{\textbf{HAFI-VLM}}, combining hierarchical, text-conditioned frequency injection with targeted adaptation of language backbone to supplement the pretrained semantic pathway with task-relevant spectral evidence. Experiments show improved fine-grained perception, text-rich understanding, and hallucination robustness across different VLM backbones. 


\bibliography{aaai2027}


\appendix

\section{Training Details}
\label{app:training_details}

\paragraph{Training Paradigm.}
Rather than training the complete model from scratch, we adopt a staged post-training procedure on top of pretrained VLM backbones. The design follows two principles. First, newly introduced frequency modules are aligned before they are jointly optimized with the visual encoder and LLM interface, preventing unstable spectral features from disrupting the pretrained vision--language representation. Second, auxiliary frequency objectives are used only when they help stabilize the frequency branch, and are removed in the final stage so that the model is optimized directly for downstream multimodal reasoning.

\begin{figure}
    \centering
    \includegraphics[width=1\linewidth]{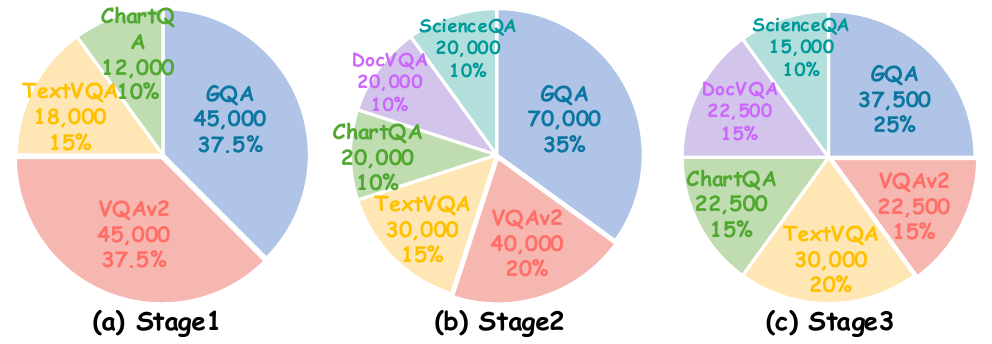}
    \caption{Stage-wise composition of the training data.}
    \label{fig:data_mixture}
\end{figure}

Let $\mathcal{L}_{\mathrm{task}}$ denote the standard autoregressive language modeling objective. We additionally use a DCT-space reconstruction loss $\mathcal{L}^{\mathrm{dct}}_{\mathrm{rec}}$ to stabilize the frequency branch and a gate-balancing loss $\mathcal{L}_{\mathrm{bal}}$ to prevent the text-conditioned band gate from collapsing to a single frequency band. The overall training data is composed of VQA, OCR, and knowledge-intensive examples; the detailed mixture schedule is visualized in Fig.~\ref{fig:data_mixture}.

\paragraph{Stage 0: Frequency Branch Pretraining.}
The first stage initializes the DCT frequency branch before it is connected to the frozen VLM. We train only the frequency encoder and the DCT reconstruction head, while the visual encoder, multimodal projector, LLM, HAFI modules, and text encoder remain unused or frozen. The objective is
\begin{equation}
\min_{\theta_f,\theta_{\mathrm{rec}}}
\mathbb{E}_{I\sim\mathcal{D}}
\left[
\mathcal{L}^{\mathrm{dct}}_{\mathrm{rec}}(I)
\right].
\end{equation}
This stage encourages the frequency branch to preserve low-, mid-, and high-frequency DCT coefficients in a stable token representation, so that later cross-attention modules can retrieve meaningful spectral evidence rather than noisy randomly initialized features.

\paragraph{Stage 1: HAFI Warm-up.}
After frequency-branch pretraining, we insert HAFI into the selected visual encoder layers and warm up the newly initialized frequency-injection modules. In this stage, only the HAFI stack is trainable. The visual encoder, frequency branch, multimodal projector, LLM, reconstruction head, and CLIP text encoder are kept frozen. The optimization objective is
\begin{equation}
\min_{\theta_{\mathrm{HAFI}}}
\mathbb{E}_{(I,T,Y)\sim\mathcal{D}}
\left[
\mathcal{L}_{\mathrm{task}}(I,T,Y)
\right].
\end{equation}
This warm-up allows the cross-attention modules and text-conditioned band gates to learn how to retrieve frequency tokens without immediately changing the pretrained visual or language representations.

\paragraph{Stage 2: Joint Frequency Adaptation.}
Stage 2 performs the main adaptation of HAFI-VLM. We jointly optimize the HAFI modules, the frequency branch, selected self-attention projections in the visual encoder, and LoRA adapters in the shallow LLM visual-enrichment layers. The multimodal projector, CLIP text encoder, visual feed-forward networks, visual LayerNorms, and deeper LLM layers remain frozen. The training objective is
\begin{equation}
\mathcal{L}^{(2)}
=
\mathcal{L}_{\mathrm{task}}
+
\beta_{\mathrm{rec}}
\mathcal{L}^{\mathrm{dct}}_{\mathrm{rec}}
+
\beta_{\mathrm{bal}}\rho(t)\mathcal{L}_{\mathrm{bal}},
\end{equation}
where $\beta_{\mathrm{rec}}=0.1$, $\beta_{\mathrm{bal}}=0.01$, and $\rho(t)$ linearly increases from $0$ to $1$ during the final portion of this stage. Delaying the gate-balancing term allows the model to first discover useful query-dependent frequency preferences before applying a weak global regularizer against degenerate band usage.

For the LLM, LoRA is applied only to the query and key projections of the shallow visual-enrichment layers. This targeted adaptation recalibrates visual--language matching for frequency-enriched visual tokens while preserving the value projections, feed-forward networks, and deeper language-reasoning layers.

\paragraph{Stage 3: Task Fine-tuning.}
The final stage removes all auxiliary objectives and fine-tunes the task pathway with
\begin{equation}
\min_{\theta_{\mathrm{HAFI}},\theta_{\mathrm{LoRA}}}
\mathbb{E}_{(I,T,Y)\sim\mathcal{D}}
\left[
\mathcal{L}_{\mathrm{task}}(I,T,Y)
\right].
\end{equation}
Only HAFI and the shallow LLM LoRA parameters are updated. The frequency branch and visual encoder are frozen. This stage aligns training with inference, where no reconstruction or gate-balancing losses are available, and consolidates the frequency-enhanced visual representation for downstream VQA-style decoding.

\paragraph{Optimization Details.}
We instantiate HAFI-VLM on LLaVA-1.5-7B and Qwen2.5-VL-7B. For LLaVA-1.5-7B, HAFI is inserted at visual layers $\{6,12,18\}$; for Qwen2.5-VL-7B, it is inserted at layers $\{8,16,24\}$. Both variants use the same DCT frequency-tokenization design and a frozen CLIP ViT-L/14 text encoder for band gating.

For LLaVA-1.5-7B, Stage 0 is trained for 8K steps with a batch size of 32 and learning rate $1\times10^{-4}$. Stage 1 is trained for 15K steps with an effective batch size of 32 and learning rate $1\times10^{-4}$. Stage 2 is trained for 25K steps with an effective batch size of 16. We use learning rates $3\times10^{-5}$ for HAFI and the frequency branch, $1\times10^{-5}$ for the selected visual self-attention projections, and $5\times10^{-6}$ for the shallow LLM LoRA parameters. Stage 3 is trained for 120K steps with an effective batch size of 16 and learning rate $1\times10^{-5}$.

For Qwen2.5-VL-7B, we follow the same staged schedule and objective design, with backbone-specific input resolution and batch sizes. Stage 0 trains the frequency branch for 8K steps. Stage 1 trains HAFI for 15K steps. Stage 2 performs joint frequency adaptation for 25K steps. Stage 3 conducts task fine-tuning for 120K steps with auxiliary losses disabled.

All experiments use AdamW, bf16 mixed precision, PyTorch, Transformers, and Hugging Face Accelerate. Unless otherwise specified, ablations are conducted from the Stage-2 checkpoint by masking selected frequency bands or injection layers at inference time, or by retraining the corresponding variant under the same protocol.

\paragraph{Datasets.}
We construct stage-specific training mixtures from GQA and VQAv2
for general visual reasoning, TextVQA, DocVQA, and ChartQA for
text-rich and OCR-oriented understanding, and ScienceQA for
knowledge-intensive reasoning. All examples are converted into a
unified instruction-following format, and only the official training
splits are used, with evaluation splits strictly excluded from
training. As illustrated in Fig.~\ref{fig:data_mixture}, the later
stages progressively increase the proportion of text-rich and
knowledge-intensive data while retaining general VQA examples to
preserve broad visual reasoning ability.

\begin{figure*}
    \centering
    \includegraphics[width=0.8\linewidth]{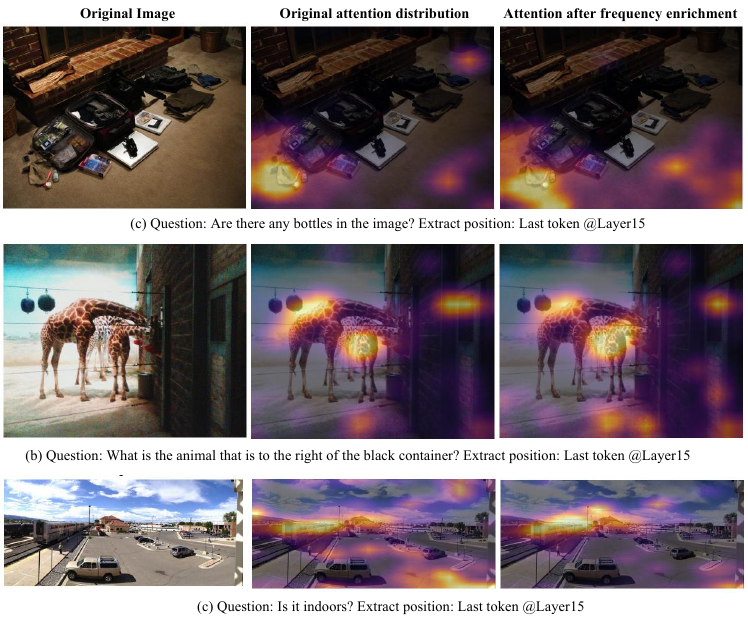}
    \caption{\textbf{Preservation of semantic grounding under frequency enrichment.} Each row presents the original image and the visual attention distributions of the base model and HAFI-VLM under the same query. Attention is extracted from the last input token immediately before answer generation at LLM Layer~15, providing a fixed middle-to-deep-layer view of the visual evidence used for decoding. Despite the redistribution of spectral information introduced by HAFI, the enhanced model retains the principal query-relevant regions attended to by the pretrained model, indicating that frequency enrichment preserves its original semantic grounding capability.}
\label{fig:attention_comparison}
\end{figure*}

\section{Preservation of semantic grounding}
Figure~\ref{fig:attention_comparison} examines whether frequency enrichment disrupts the semantic grounding capability established by the pretrained VLM. We extract visual attention from the last input token immediately before autoregressive decoding at Layer~15, a representative middle-to-deep layer where visual evidence has been integrated with the query but answer generation has not yet begun. We use a fixed layer rather than averaging over the full language backbone because shallow and deep layers encode different stages of visual--language interaction, and their aggregation may obscure the semantic structure present at a particular depth; cross-layer behavior is analyzed separately through attention rollout and is not illustrated. Across object-presence, spatial-relation, and scene-level queries, HAFI-VLM largely preserves the dominant spatial regions attended to by the base model. The observed differences mainly reflect local redistribution within or around these semantically relevant regions, rather than a displacement of attention toward unrelated structures. Together with the broader task-dependent frequency distributions illustrated in the main paper, these results show that HAFI expands the spectral evidence available to the model without overwriting the pretrained semantic grounding pathway.

\end{document}